\ifdefined\pdfoutput\pdfoutput=1\fi
\documentclass[10pt, a4paper]{article}

\usepackage[final]{lrec2026}
\usepackage{latexsym}
\usepackage{adjustbox}

\usepackage[margin=1in]{geometry}

\usepackage{microtype}

\usepackage{float}
\usepackage{inconsolata}

\usepackage{graphicx}

\usepackage{booktabs}

\usepackage{xcolor}

\usepackage{multirow}

\usepackage{xcolor}
\usepackage{pifont}

\newcommand{\greentick}{\textcolor{green}{\ding{52}}}
\newcommand{\graytick}{\textcolor{gray}{\ding{52}}}
\newcommand{\redcross}{\textcolor{red}{\ding{55}}}

\title{\emph{OasisSimp}: An Open-source Asian-English Sentence Simplification Dataset}

\name{
Hannah Liu$^{1\ast}$, Muxin Tian$^{1\ast}$, Iqra Ali$^{2}$, Haonan Gao$^{3}$, Qiaoyiwen Wu$^{1}$,\\
{\bf \large Blair Yang$^{4}$, Uthayasanker Thayasivam$^{5}$, En-Shiun Annie Lee$^{6,1}$,}\\
{\bf \large Pakawat Nakwijit$^{2}$, Surangika Ranathunga$^{7}$, Ravi Shekhar$^{8}$}
}

\address{
$^{1}$University of Toronto, $^{2}$Queen Mary University of London, $^{3}$Yale University, $^{4}$CoolWei AI Lab\\
$^{5}$University of Moratuwa, $^{6}$Ontario Tech University, $^{7}$Massey University, $^{8}$University of Essex\\
\{hannahhere.liu, murphy.tian, qiaoyiwen.wu\}@mail.utoronto.ca, \\ \{iqra.ali, p.nakwijit\}@qmul.ac.uk,  
eric.gao@yale.edu, yangliwei@coolwei.com, \\ rtuthaya@cse.mrt.ac.lk, 
annie.lee@ontariotechu.ca, \\ s.ranathunga@massey.ac.nz, r.shekhar@essex.ac.uk
}

\abstract{
Sentence simplification aims to make complex text more accessible by reducing linguistic complexity while preserving the original meaning. However, progress in this area remains limited for mid-resource and low-resource languages due to the scarcity of high-quality data. To address this gap, we introduce the \emph{OasisSimp} dataset, a multilingual dataset for sentence-level simplification covering five languages: English, Sinhala, Tamil, Pashto, and Thai. Among these, no prior sentence simplification datasets exist for Thai, Pashto, and Tamil, while limited data is available for Sinhala. Each language simplification dataset was created by trained annotators who followed detailed guidelines to simplify sentences while maintaining meaning, fluency, and grammatical correctness. We evaluate eight open-weight multilingual Large Language Models (LLMs) on the OasisSimp dataset and observe substantial performance disparities between high-resource and low-resource languages, highlighting the simplification challenges in multilingual settings. The OasisSimp dataset thus provides both a valuable multilingual resource and a challenging benchmark, revealing the limitations of current LLM-based simplification methods and paving the way for future research in low-resource sentence simplification. The dataset is available at \url{https://OasisSimpDataset.github.io/}.
 \\ \newline  }

\begin{document}

\maketitleabstract

\renewcommand{\thefootnote}{\fnsymbol{footnote}}
\footnotetext[1]{These authors contributed equally.}
\renewcommand{\thefootnote}{\arabic{footnote}}

\section{Introduction}
\label{sec:introduction}

Reading is a critical daily activity in our text-rich society, yet it can be challenging for some due to complex words, sentences, or meanings. Organizations such as the Plain Language Association International (PLAIN)\footnote{\url{https://plainlanguagenetwork.org/}} and the Easy-to-Read movement~\citep{de2018lectura} emphasize using simple texts to make information accessible to all. Ensuring written content is clear and readable is crucial for equitable knowledge access. 

Sentence simplification is the task of transforming complex text into an easier-to-understand format by reducing its linguistic complexity while preserving the original meaning and factual information. This process is essential for enhancing information accessibility for a wide range of individuals, including language learners, people with cognitive or reading disabilities such as dyslexia or aphasia, and younger audiences~\citep{paetzold2017survey, espinosa-zaragoza-etal-2023-review, al2021automated}. High-quality simplified text can also be a valuable preprocessing step for various downstream Natural Language Processing (NLP) tasks, including Machine Translation, text summarization, and question answering \citep{dadu2021text}. 

In the last two decades, the NLP community has made significant progress in addressing this challenge by developing various datasets and tools for sentence simplification. Early efforts primarily focused on high-resource languages such as English~\citep{shardlow-2014-survey, alva-manchego-etal-2020-survey} and Spanish~\citep{ferres-saggion-2022-alexsis}. In particular, the availability of English datasets such as Newsela~\citep{xu-etal-2015-problems} and WikiLarge~\citep{zhang-lapata-2017-sentence} has driven substantial advances in simplification model development. However, this focus has created a pronounced resource imbalance, leaving many other languages, especially low-resource languages such as Sinhala and Pashto, significantly underserved. The scarcity of high-quality, parallel complex-simple sentence pairs for these languages presents a major obstacle to developing, evaluating, and advancing simplification technologies across linguistic boundaries.

Recent efforts have sought to expand language coverage. For example, \citet{ryan-etal-2023-revisiting} introduced the MULTISIM benchmark, covering 12 languages using data from 27 sources. Similarly, multilingual datasets for domain-specific applications, particularly in the medical domain~\citep{basu2023med}, have emerged, including {M}ulti{C}ochrane~\citep{joseph-etal-2023-multilingual} and {M}ulti{MSD}~\citep{horiguchi-etal-2025-multimsd}. However, many of these datasets are domain-specific and offer limited general language representation.

We introduce the \emph{OasisSimp} dataset (An \underline{O}pen-source \underline{Asi}an-English \underline{S}entence \underline{Simp}lification Dataset), a publicly available dataset for sentence-level simplification to address this critical resource gap. OasisSimp covers five languages - English, Sinhala, Tamil, Thai, and Pashto - with all parallel sentence pairs created through expert human simplification. Trained annotators followed consistent, detailed guidelines to manually rewrite complex sentences into simpler forms, ensuring that the core meaning, grammatical correctness, and fluency were preserved.
Beyond releasing this resource, we leverage OasisSimp to evaluate LLM-based text simplification approaches proposed in ~\citet{kew-etal-2023-bless}. We report performance of various various LLMs in zero- and few-shot settings and report SARI~\citep{xu-etal-2016-optimizing} and BERTScore~\citep{zhang2020bertscoreevaluatingtextgeneration}.

By providing both the \emph{OasisSimp} dataset and associated benchmarking results, we aim to expand the resources available for multilingual sentence simplification research significantly. We anticipate that this work will stimulate further research, support the development of more inclusive and effective simplification systems, and ultimately make information more accessible to diverse global audiences.

\section{Related Work}

Sentence Simplification makes written language easier to read and understand while keeping the original meaning intact. It plays an important role in improving the readability and accessibility of text for diverse audiences. Although research in sentence simplification has a long and rich history, it has mainly focused on high-resource languages, particularly English. Simplification generally occurs at three levels: lexical, syntactic, and conceptual or discourse. Lexical simplification involves replacing complex words with simpler alternatives~\citep{venugopal-etal-2022-cwid}, while syntactic simplification reduces grammatical complexity to make sentences clearer and easier to follow~\citep{shardlow-2014-survey,alva-manchego-etal-2020-survey}. On the other hand, conceptual simplification aims to make the underlying ideas easier to understand, not just the words or sentence structures used to express them~\citep{eschenbrucher2021makes}. Most studies have concentrated on lexical and syntactic simplification, and this work focuses explicitly on sentence-level syntactic simplification.

\paragraph{Sentence Simplification Datasets} 

Early and influential datasets for English sentence simplification are Simple English Wikipedia \citep{coster-kauchak-2011-simple}, which was constructed by automatically aligning complex sentences with their simplified counterparts, with previous research on automatically constructing parallel data \citep{kajiwara-komachi-2016-building}. Other prominent English simplification datasets include ASSET \citep{alva-manchego-etal-2020-asset}, developed through crowdsourcing, and WikiLarge \citep{zhang-lapata-2017-sentence}, created by combining multiple existing resources. These corpora have been instrumental in advancing the field from early rule-based and statistical methods to more recent neural approaches to sentence simplification (e.g., \citealp{nisioi-etal-2017-exploring}; \citealp{martin-etal-2020-controllable}).

In recent years, research in sentence simplification has gradually expanded beyond English to include other languages. For Italian, \citet{tonelli2016simpitiki} and \citet{brunato-etal-2016-paccss} introduced sentence-level simplification datasets, with the former focusing on manual simplification and the latter employing automatic methods. For Danish, the DSim corpus \citep{klerke-sogaard-2012-dsim} was created using journalists producing simplifications, while \citet{grabar-cardon-2018-clear} developed a French medical simplification dataset combining crowdsourced and expert annotations. For Russian, the RuSimpleSentEval-2021 Shared Task \citep{sakhovskiy2021rusimplesenteval} provided a crowdsourced dataset with both automatic and manual sentence alignments. A notable advancement in multilingual simplification is the MultiSim benchmark \citep{stajner-etal-2022-multisim}, which includes English, Spanish, and Portuguese data. Although this benchmark marked an important step towards multilingual evaluation, it remains centered on high-resource languages. Similarly, \citet{ryan-etal-2023-revisiting} proposed a non-English benchmark by combining existing resources, but coverage for genuinely low-resource languages remains limited.

The availability of sentence simplification corpora for low-resource languages is still scarce. The SiTSE corpus \citep{ranathunga2025sitse} includes 1,000 complex Sinhala sentences simplified by three annotators, yielding 3,000 complex–simple pairs. Likewise, \citet{mondal2024dimsim} created Bengali and Marathi datasets, each containing 500 pairs, while \citet{anees2021automatic} developed an Urdu corpus comprising 610 sentences simplified by two annotators. Although these initiatives represent valuable progress, they are relatively small compared to English datasets. To address this gap, our work introduces the OasisSimp dataset, a new multilingual sentence simplification dataset covering mid- and low-resource languages.

\paragraph{Simplification Methodology} 

Supervised learning has long been the dominant paradigm in modern sentence simplification~\citep{al2021automated,alva-manchego-etal-2020-data}. However, supervised approaches are often impractical for low-resource languages, where parallel simplification training data is limited or unavailable. To address this issue, researchers have often turned to machine translation (MT) to create pseudo data, for example, by translating complex English sentences into another language and simplifying them, or by translating simplified English sentences into the target language \cite{palmero-aprosio-etal-2019-neural, qiang-etal-2021-unsupervised-crosslingual, sheang-etal-2021-multilingual, ki-carpuat-2025-automatic}. Although these methods are creative, they can introduce translation noise and fail to capture language-specific nuances of simplification \cite{hasan-etal-2021-introducing}. Further improvements were achieved by mining paraphrases from large-scale web-crawled data~\citep{martin-etal-2022-muss}. More recently, the emergence of LLMs has enabled simplification through zero-shot and few-shot prompting, reducing reliance on explicit training data. In this work, we adopt an LLM-based simplification approach, leveraging the strong generalization capabilities of multilingual LLMs to handle sentence simplification across diverse languages.

\section{Data Creation}
\label{sec:dataset_creation}

\subsection{Language Selection}
\label{ssec:language_selection}

We selected five diverse languages (English, Thai, Tamil, Sinhala, and Pashto)\footnote{Based on data availability, English: 5, Thai \& Tamil: 3, Sinhala \& Pashto: 2~\citep{joshi-etal-2020-state, ranathunga-de-silva-2022-languages}, where 5 is high-resources and 1 is low-resource language.} to create a sentence simplification dataset, based on their linguistic, typological, and sociolinguistic diversity. English, a high-resource Indo-European language, provides well-studied benchmarks for comparison. Thai, a tonal Southeast Asian language with a script that lacks spaces, Tamil, a morphologically rich Dravidian language, Sinhala, an Indo-Aryan language with complex syllable structures, and Pashto, an Eastern Iranian language with dialectal variations, are low-resource languages with diverse scripts and grammatical features. This selection covers multiple language families, scripts (Latin, Thai, Brahmi-derived, Arabic-derived), and linguistic characteristics (analytic vs. agglutinative, tonal vs. non-tonal). Apart from that, Thai, Tamil, and Pashto have no sentence simplification datasets. For Sinhala, only a small dataset exists~\citep{ranathunga2025sitse}.  Including these languages addresses a significant gap in sentence simplification research. Additionally, all five languages are spoken by sizable populations, meaning that simplification can have a real social impact by improving literacy, education, and access to information. The resulting dataset will support multilingual research and practical applications across languages with very different linguistic and writing systems.

\subsection{Data Source and Selection}
\label{ssec:data_source}

Due to the lack of a common data source across all five languages, we individually selected sources for each language, ensuring they are publicly available and cover diverse domains such as Wikipedia, newspapers, and government documents. Our goal was to collect sentences with varying lengths and complexity. We applied automatic filtering criteria on the selected source data, including a minimum sentence length. We also ensured coverage across different topics. For example, Wikipedia articles on animals, products, buildings, etc., and newspaper articles on sports, politics, entertainment, and more. In addition to these general steps, we applied language-specific filtering to account for linguistic characteristics unique to each language (see below). This approach provides a wide range of sentence complexity. Finally, after automatically identifying a large set of potential complex sentences, we manually reviewed them to select a number for simplification. Manual inspection ensured the chosen sentences were genuinely complex and suitable for simplification.

\paragraph{English (\emph{OasisSimp-EN})}
The complex English sentences were drawn from Canadian general-interest newspapers via the NewsEdits curation workflow, primarily \textit{The Globe and Mail}. We sampled across general-interest sections (e.g., \textit{News}, \textit{Report on Business}, \textit{Opinion}, \textit{Arts \& Life/Books}) and excluded templated material such as stock tables, listings, photo captions, corrections, and headlines. Candidate sentences were restricted to 100--300 characters and excluded if they contained special characters (e.g., ``@'', ``\&''). We also limited proper nouns to reduce world-knowledge dependency and annotation ambiguity, thereby encouraging structural/lexical simplification over entity substitution. To maximize sentence‑level simplifiability, we prioritized sentences whose complexity stems from syntax (e.g., coordination and subordination, apposition, heavy noun phrases) and lexical density rather than from named‑entity load. This filtering produced a large pool of candidate complex sentences, from which we manually curated 2{,}500 context‑independent complex sentences for simplification to form the \emph{OasisSimp-EN}.

\paragraph{Sinhala (\emph{OasisSimp-SI}) \& Tamil (\emph{OasisSimp-TA})} Sinhala and Tamil complex sentences were selected from the SiTa trilingual parallel corpus \citep{DBLP:journals/corr/abs-2011-02821, ranathunga2018si}. This corpus has about 50k unique sentence pairs coming from official government documents, meticulously cleaned and fixed by language experts. These government documents are more complex than other sources from Sri Lanka, such as news or social media text. Government documents often feature longer sentences, domain-specific terminology from fields such as accounting and law, and a highly formal register~\cite{ranathunga2025sitse}. We opted to select sentences containing rare words. Initially, rare word based filtering was done for Sinhala (we selected rare words that have a frequency of less than 5-50\footnote{words that had a frequency less than 5 were mostly misspelled words.} in 140k sentences of the common crawl corpus). This resulted in 5859 sentences. Manual observation revealed that some sentences are just lists
of items, and some are near duplicates. Therefore, these sentences were manually
filtered. Finally, we selected 2500 long sentences. In certain cases, there were technical terms, and within brackets, the English
meaning was given. These English terms were removed. A random 500 were selected out of these sentences, and the corresponding Tamil sentences were retrieved from the SiTa trilingual corpus. The Tamil dataset had to be restricted to 520 due to the unavailability of human annotators. Each complex sentence was simplified by 5 annotators, for both Sinhala and Tamil, to form the \emph{OasisSimp-SI} and \emph{OasisSimp-TA} datasets.

\paragraph{Thai (\emph{OasisSimp-TH})} We used the ThaiSum dataset as our initial corpus \citep{chumpolsathien_2020}. This dataset was collected from Thai news websites: \hyperlink{https://www.thairath.co.th/home}{Thairath}, \hyperlink{https://www.thaipbs.or.th/home}{ThaiPBS}, \hyperlink{https://prachatai.com/}{Prachatai}, and \hyperlink{https://thestandard.co/homepage/}{The Standard}. A Thai-specific CRF segmentator trained on the TED dataset \citep{pythainlp} pre-segmented each document into sentences. We first randomly sampled sentences from the corpus and performed a filtering based on the rare words, length, and topic diversity. We selected sentences containing rare and non-rare words to avoid rare word bias. All sentences were also subjected to a length constraint, with only those more than 15 words being retained. To ensure topic diversity, we re-sampled the pre-filtered sentences from 10 distinct news categories: Politics, Local News, Economics, Society, Foreign Affairs, Quality of Life, Human Rights, Lifestyle, Culture, and Education, resulting in a pool of 3,815 sentences. 
In addition, due to the ambiguity of sentence and word boundaries and the limitations of the automatic sentence segmentation, one of the authors manually picked the correct sentence boundary. We finally obtained 1499 Thai sentences to form the \emph{OasisSimp-TH} dataset.

\paragraph{Pashto (\emph{OasisSimp-PS})} We used Wikipedia as our primary source for collecting complex Pashto sentences. Wikipedia is a reliable, publicly available, and diverse domain resource for Pashto, providing naturally occurring complex sentences with authentic vocabulary, syntax, and named entities.  We initially collected 10,000 Pashto sentences, distributed equally across ten semantic categories like animals, products, buildings, locations/places, events, food, drinks, hobbies, works of art, and organizations \textit{(1,000 sentences per category)}. Human annotators manually extracted these sentences from Wikipedia to ensure relevance and domain coverage within each category. Out of the 1,000 sentences per category, the top \textit{250 }most complex sentences were selected based on four key points: sentence length, syntactic complexity, vocabulary richness, and semantic depth. Longer (25/30 words), multi-clause sentences with structured and diverse vocabulary are preferred, ensuring linguistic richness and contextual nuance. This filtering resulted in a final set of \textit{2,500} sentences and formed the \emph{OasisSimp-PS} dataset, ensuring sufficient complexity for evaluating simplification systems. All selected sentences were reviewed by a native Pashto linguistic expert to ensure their cultural and linguistic appropriateness before annotation. The evaluation mainly depends on cultural relevance, naturalness, grammatical correctness, vocabulary suitability, and semantic clarity.

\begin{table}[!htbp]
\centering
\footnotesize
\caption{Final statistics for each language in the \emph{OasisSimp} dataset. Comp - complex, Simp - simplified.}
\label{tab:dataset_stats_full}
\resizebox{\columnwidth}{!}{
\begin{tabular}{lccccc}
\toprule
\textbf{Lang} &
\shortstack{\textbf{\# Comp}\\\textbf{Sentences}} &
\shortstack{\textbf{Avg. Simp}\\\textbf{Sentences}} &
\shortstack{\textbf{Avg.Comp}\\\textbf{ Length}} &
\shortstack{\textbf{Avg. Simp}\\\textbf{ Length}} &

\shortstack{\textbf{Source}\\\textbf{Domain}} \\
\midrule
English & 2500 & 2.86 & 24.35 & 17.23  & News \\
Sinhala & 2500 & 5.00 & 30.12 & 28.78  & Govt \\
Thai & 1499 & 5.06 & 48.24 & 37.77  & News \\
Tamil & 520 & 4.66 & 23.22 & 17.65  & Govt \\
Pashto & 2500 & 3.00 & 28.81 & 20.31  & Wiki \\
\bottomrule
\end{tabular}
}

\end{table}

\subsection{Data Annotation Workflow}
\label{ssec:human_simplification_en}

\begin{table*}[htb!]
\centering
\caption{Sample Examples from the \emph{OasisSimp} Dataset for each language.}
\label{tab:dataexample}
\begin{tabular}{c}
\includegraphics[width=0.97\textwidth]{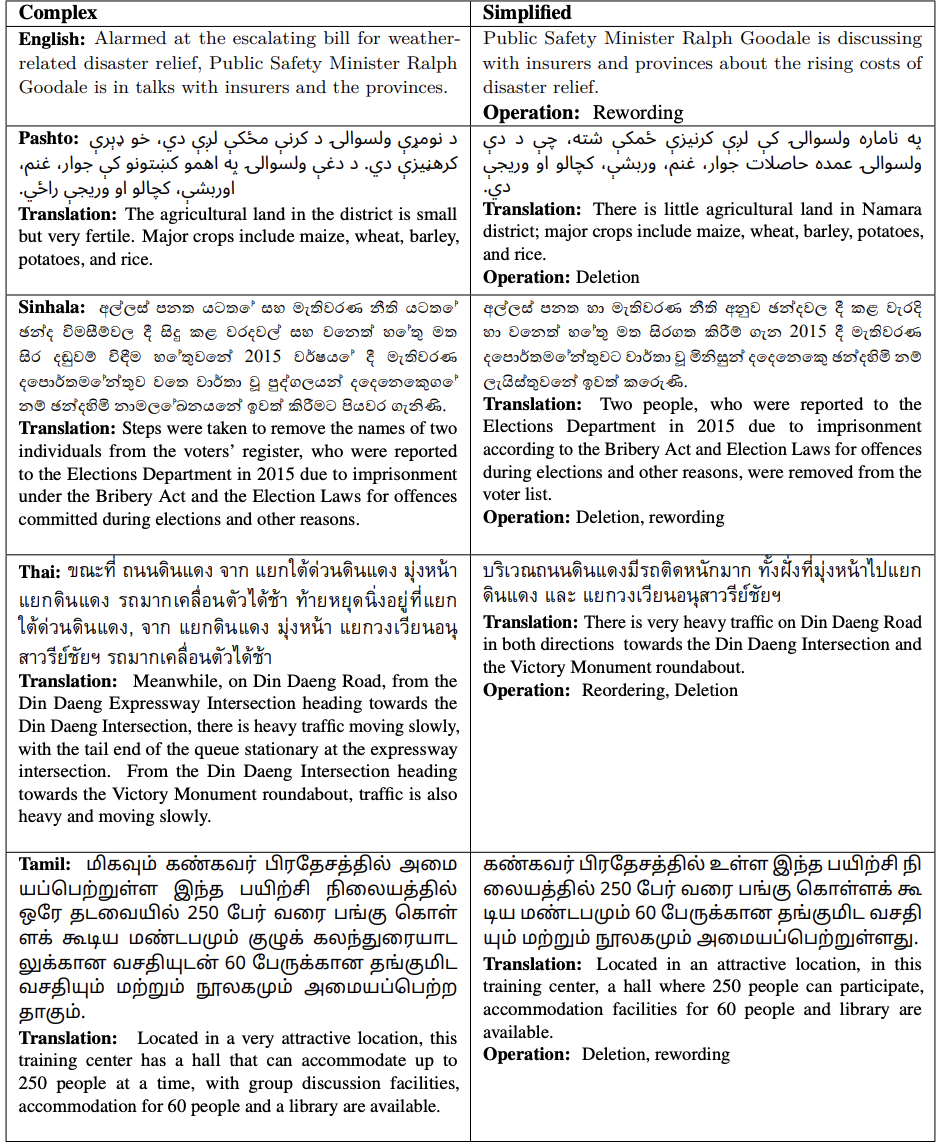} \\
\end{tabular}

\end{table*}

We used 3-6 native language speakers with at least Bachelor’s degrees for annotation. We recruited the annotators based on our contacts from the respective countries to have greater quality control over the annotation process, instead of going for an online platform. We instructed annotators to simplify sentences while preserving meaning, fluency, and grammaticality. We provided detailed instructions, adapted from ~\citet{xu-etal-2016-optimizing, alva-manchego-etal-2020-asset}. Before annotation, annotators received structured training sessions to familiarize them with the simplification guidelines, operational definitions, and annotation criteria. During the training phase, annotators were encouraged to ask questions to ensure they understood how to retain the core meaning while making sentences easier for lower-literacy speakers. We provided at least 3 rounds of training to each annotator. During training, annotators were provided with illustrative examples to demonstrate the following key simplification operations. 

\noindent \textbf{-} Replace rare or technical terms with simpler synonyms \textit{(Rewording)}. \\
\noindent \textbf{-} Split long sentences into multiple shorter sentences \textit{(Sentence Splitting)}. \\
\noindent \textbf{-} Remove unnecessary details while preserving meaning \textit{(Deletion)}. \\
\noindent \textbf{-} Restructure sentences for clarity \textit{(Reordering)}.

The annotations were performed in batches of 25-100 sentences, depending on language, complexity, and the annotator's capacity. Initially, one of the authors verified that the annotators followed the instructions. In case of any issue, we provided further instructions to the annotator. After a few rounds of building confidence with the annotator, we randomly sampled some simplified text for verification. In Table~\ref{tab:dataexample}, we present one example from each language with the corresponding transaction and operation performed.

Table~\ref{tab:dataset_stats_full} presents the final statistics for each language in the OasisSimp dataset, including the number of complex sentences, the average number of simplified sentences per complex instance, the average sentence lengths for both complex and simplified versions, and their respective sources. The sentence length for Thai is measured by tokens, while for other languages it's measured by the number of words.  In most languages, each complex sentence is paired with three simplified versions, except for a few English sentences that were excluded due to quality issues. As shown across all languages, the average length of simplified sentences is consistently shorter than that of complex sentences, indicating that simplification often involves the removal of redundant or less essential information. We divided the dataset into validation and test subsets to support unsupervised model development, allocating 80\% of the data to testing and the remaining 20\% to validation.

\section{Methodology and Evaluation}
\label{sec:evaluation_methodology}

Our work focuses on advancing less-resource sentence simplification datasets, rather than proposing a new simplification methodology. To this end, we adopt an existing LLM-based simplification framework, following the prompting strategy proposed by \citet{kew-etal-2023-bless}, where the model is instructed to ``Simplify the given sentence …'' to generate simplified outputs. This approach allows us to systematically evaluate the capabilities of different multilingual models under a consistent simplification paradigm, isolating the effect of model design and linguistic coverage on performance.

We evaluated a diverse set of open-weight, multilingual LLMs, including Aya (Aya-Expanse-8B), Cmd-R (c4ai-command-r7b-12-2024), DeepSeek (deepseek-llm-7B-chat), EuroLLM (EuroLLM-9B-Instruct), Gemma (Gemma-3-12B-it), LLaMA (Llama-3.2-3B-Instruct), Mistral (Mistral-7B-Instruct-v0.2), and Qwen (Qwen2.5-7B-Instruct). These models encompass a range of architectural designs, parameter scales, and training paradigms. While not all models explicitly disclose the full list of languages in their training data, we broadly categorize them based on whether a language is known to be included, excluded, or likely included based on an educated guess (e.g., Gemma). We also include models such as EuroLLM, trained primarily on European languages, to examine their ability to generalize to unseen or low-resource languages. Based on the available language information, all models include English. LLaMA includes Tamil and Thai, while Gemma is most likely trained on all evaluated languages except Pashto.

We conduct experiments in both zero-shot and few-shot settings, building upon BLESS~\citep{kew-etal-2023-bless}. For zero-shot, we experimented with multiple temperatures (0.1 - 0.9 in steps of 0.1) and report the best performance. In the few-shot configuration, we used 1-shot and 5-shot settings using the best temperature from zero-shot. This design enables a systematic comparison of LLMs with varying capacities and linguistic generalization abilities in the context of less-resource sentence simplification. We used the two standard automatic simplification metrics for the evaluation: SARI~\citep{xu-etal-2016-optimizing} and BERTScore~\citep{zhang2020bertscoreevaluatingtextgeneration}. SARI measures the goodness of words added, deleted, and kept by the simplification system, while BERTScore measures the semantic similarity of the simplified sentence with the reference sentences\footnote{ We didn't use BLEU Score because of its unsuitability~\citep{callison-burch-etal-2006-evaluating, reiter-2018-structured}. However, for completeness, we included BLEU results in Appendix.}.

\section{Results and Discussion}
\label{sec:experimental_results}
In Table~\ref{tab:english}, ~\ref{tab:sinhala}, ~\ref{tab:thai}, ~\ref{tab:tamil}, ~\ref{tab:pashto}, we present Zero, 1, and 5-shot SARI and BERTScore (\textbf{F}\textsubscript{ref}) using multiple LLMs from English, Sinhala, Thai, Tamil, and Pashto, respectively.

\begin{table*}[ht!]
\centering
\footnotesize
\setlength{\tabcolsep}{2pt}
\caption{\textbf{Results on English (\emph{OasisSimp-EN}) dataset}: Zero, 1- and 5-shot Results. Highest value in \textbf{bold} and lowest \underline{underlined}.  \greentick: Language included in LLM training; \graytick: Likely included (Educated Guess); \redcross: Not included.}
\resizebox{\textwidth}{!}{
\begin{tabular}{l ccccc | ccccc | ccccc}
\toprule
\multirow{3}{*}{\textbf{Model}} & \multicolumn{5}{c|}{\textbf{0 Shot}} & \multicolumn{5}{c|}{\textbf{1 Shot}} & \multicolumn{5}{c}{\textbf{5 Shot}} \\
\cmidrule(lr){2-6} \cmidrule(lr){7-11} \cmidrule(lr){12-16}
 & \multicolumn{3}{c}{\textbf{SARI Comp.}} & \multirow{2}{*}{\textbf{SARI}} & \multirow{2}{*}{\textbf{F}\textsubscript{ref}} & \multicolumn{3}{c}{\textbf{SARI Comp.}} & \multirow{2}{*}{\textbf{SARI}} & \multirow{2}{*}{\textbf{F}\textsubscript{ref}} & \multicolumn{3}{c}{\textbf{SARI Comp.}} & \multirow{2}{*}{\textbf{SARI}} & \multirow{2}{*}{\textbf{F}\textsubscript{ref}} \\
\cmidrule(lr){2-4} \cmidrule(lr){7-9} \cmidrule(lr){12-14}
 & \textbf{ADD} & \textbf{KEEP} & \textbf{DEL} & & & \textbf{ADD} & \textbf{KEEP} & \textbf{DEL} & & & \textbf{ADD} & \textbf{KEEP} & \textbf{DEL} & & \\
\midrule
Aya \greentick & 9.32 & 44.98 & 75.23 & 43.18 & 54.44 & 9.68 & 44.90 & 72.51 & 42.36 & 56.35 & 10.18 & 45.91 & 71.16 & \underline{42.42} & 57.20 \\
Cmd-R \greentick & \textbf{9.69} & 44.95 & 72.89 & 42.51 & 55.90 & \textbf{10.99} & 43.71 & 77.57 & 44.09 & 55.03 & \textbf{11.91} & 45.28 & 77.09 & 44.76 & 56.63 \\
DeepSeek \greentick & 7.03 & \underline{41.47} & 76.30 & 41.60 & 51.88 & 7.80 & \underline{41.12} & 76.82 & 41.91 & \underline{51.92} & 9.41 & \underline{42.03} & 77.22 & 42.89 & \underline{54.15} \\
EuroLLM \greentick & 9.32 & 45.60 & 68.36 & 41.10 & \textbf{56.98} & \textbf{10.99} & \textbf{46.98} & \underline{69.35} & 42.44 & \textbf{57.96} & 11.63 & 46.55 & \underline{70.93} & 43.04 & \textbf{58.10} \\
Gemma \greentick & \underline{5.24} & 44.43 & 68.54 & 39.40 & 51.87 & \underline{6.55} & 43.26 & 74.44 & \underline{41.41} & 52.34 & \underline{9.19} & 44.67 & 77.06 & 43.64 & 55.27 \\
LLaMA & 6.48 & 43.31 & \underline{68.34} & \underline{39.38} & 54.30 & 8.11 & 43.42 & 72.83 & 41.45 & 54.53 & 9.93 & 44.75 & 73.75 & 42.81 & 56.00 \\
Mistral \greentick & 8.56 & 43.66 & \textbf{77.46} & \textbf{43.23} & 52.49 & 10.31 & 43.82 & \textbf{78.43} & 44.18 & 54.55 & 11.61 & 44.01 & \textbf{78.59} & 44.74 & 55.89 \\
Qwen \greentick & 8.70 & \textbf{46.07} & 73.53 & 42.77 & \underline{42.36} & 9.54 & 46.40 & 77.25 & \textbf{44.39} & 53.03 & 10.88 & \textbf{47.01} & 77.08 & \textbf{44.99} & 55.27 \\
\bottomrule
\end{tabular}
}
\label{tab:english}
\end{table*}

\begin{table*}[ht!]
\centering
\footnotesize
\setlength{\tabcolsep}{2pt}
\caption{\textbf{Results on Sinhala (\emph{OasisSimp-SI}) dataset}: Zero, 1- and 5-shot Results.  Highest value in \textbf{bold} and lowest in \underline{underlined}.  \greentick: Language included in LLM training; \graytick: Likely included (Educated Guess); \redcross: Not included.}
\label{tab:sinhala_formatted}

\resizebox{\textwidth}{!}{
\begin{tabular}{l ccccc | ccccc | ccccc}
\toprule
\multirow{3}{*}{\textbf{Model}} & \multicolumn{5}{c|}{\textbf{0 Shot}} & \multicolumn{5}{c|}{\textbf{1 Shot}} & \multicolumn{5}{c}{\textbf{5 Shot}} \\
\cmidrule(lr){2-6} \cmidrule(lr){7-11} \cmidrule(lr){12-16}
& \multicolumn{3}{c}{\textbf{SARI Comp.}} & \multirow{2}{*}{\textbf{SARI}} & \multirow{2}{*}{\textbf{F}\textsubscript{ref}} & \multicolumn{3}{c}{\textbf{SARI Comp.}} & \multirow{2}{*}{\textbf{SARI}} & \multirow{2}{*}{\textbf{F}\textsubscript{ref}} & \multicolumn{3}{c}{\textbf{SARI Comp.}} & \multirow{2}{*}{\textbf{SARI}} & \multirow{2}{*}{\textbf{F}\textsubscript{ref}} \\
\cmidrule(lr){2-4} \cmidrule(lr){7-9} \cmidrule(lr){12-14}
& \textbf{ADD} & \textbf{KEEP} & \textbf{DEL} & & & \textbf{ADD} & \textbf{KEEP} & \textbf{DEL} & & & \textbf{ADD} & \textbf{KEEP} & \textbf{DEL} & & \\
\midrule
Aya \redcross & 0.47 & 23.11 & 68.49 & 30.69 & 54.94 & 0.40 & 26.01 & 67.39 & 31.27 & 54.22 & 0.21 & 27.62 & 65.75 & 31.20 & 44.78 \\
Cmd-R \redcross & 0.18 & \textbf{29.95} & \underline{65.05} & 31.73 & 58.32 & 0.19 & 30.24 & \underline{65.08} & 31.84 & 55.94 & 0.13 & 28.87 & \underline{65.18} & 31.40 & 45.34 \\
DeepSeek \redcross & 0.16 & 16.92 & 68.71 & 28.60 & \underline{-0.14} & 0.14 & 24.99 & 66.71 & 30.61 & 57.68 & 0.08 & 20.23 & 67.60 & 29.30 & 38.55 \\
EuroLLM \redcross & \underline{0.03} & 19.19 & 68.14 & 29.12 & 47.59 & \underline{0.04} & \underline{18.74} & 68.18 & \underline{28.99} & 47.56 & \underline{0.00} & \underline{2.03} & \textbf{70.18} & \underline{24.07} & \underline{-60.47} \\
Gemma \graytick & \textbf{3.89} & 28.64 & \textbf{71.77} & \textbf{34.77} & \textbf{66.77} & \textbf{4.62} & \textbf{36.89} & \textbf{71.25} & \textbf{37.59} & \textbf{70.44} & \textbf{5.65} & \textbf{44.21} & 69.82 & \textbf{39.89} & \textbf{73.89} \\
LLaMA \redcross & 0.32 & 17.77 & 69.25 & 29.11 & 47.10 & 0.31 & 19.23 & 68.91 & 29.49 & \underline{45.49} & 0.30 & 20.49 & 68.18 & 29.66 & 32.99 \\
Mistral \redcross & 0.19 & \underline{11.61} & 70.00 & \underline{27.26} & 42.84 & 0.22 & 19.58 & 68.60 & 29.47 & 54.85 & 0.17 & 25.28 & 66.80 & 30.75 & 58.47 \\
Qwen \redcross & 0.46 & 28.70 & 66.27 & 31.81 & 59.50 & 0.43 & 29.17 & 65.84 & 31.81 & 57.44 & 0.38 & 29.01 & 65.66 & 31.68 & 46.73 \\
\bottomrule
\end{tabular}
}
\label{tab:sinhala}
\end{table*}

\begin{table*}[ht!]
\centering
\footnotesize
\setlength{\tabcolsep}{2pt}
\caption{\textbf{Results on Thai (\emph{OasisSimp-TH}) dataset}: Zero, 1- and 5-shot Results. Highest value in \textbf{bold} and lowest in \underline{underlined}.  \greentick: Language included in LLM training; \graytick: Likely included (Educated Guess); \redcross: Not included.}

\resizebox{\textwidth}{!}{
\begin{tabular}{l ccccc | ccccc | ccccc}
\toprule
\multirow{3}{*}{\textbf{Model}} & \multicolumn{5}{c|}{\textbf{0 Shot}} & \multicolumn{5}{c|}{\textbf{1 Shot}} & \multicolumn{5}{c}{\textbf{5 Shot}} \\
\cmidrule(lr){2-6} \cmidrule(lr){7-11} \cmidrule(lr){12-16}
& \multicolumn{3}{c}{\textbf{SARI Comp.}} & \multirow{2}{*}{\textbf{SARI}} & \multirow{2}{*}{\textbf{F}\textsubscript{ref}} & \multicolumn{3}{c}{\textbf{SARI Comp.}} & \multirow{2}{*}{\textbf{SARI}} & \multirow{2}{*}{\textbf{F}\textsubscript{ref}} & \multicolumn{3}{c}{\textbf{SARI Comp.}} & \multirow{2}{*}{\textbf{SARI}} & \multirow{2}{*}{\textbf{F}\textsubscript{ref}} \\
\cmidrule(lr){2-4} \cmidrule(lr){7-9} \cmidrule(lr){12-14}
& \textbf{ADD} & \textbf{KEEP} & \textbf{DEL} & & & \textbf{ADD} & \textbf{KEEP} & \textbf{DEL} & & & \textbf{ADD} & \textbf{KEEP} & \textbf{DEL} & & \\
\midrule
Aya \redcross & 0.33 & 21.01 & 84.30 & 35.22 & 56.49 & 0.38 & 24.45 & 83.53 & 36.12 & 59.01 & 0.43 & 25.98 & 82.59 & 36.33 & 60.66 \\
Cmd-R \redcross & 0.47 & 29.02 & 82.19 & 37.23 & \textbf{62.17} & 0.51 & 26.88 & 81.85 & 36.41 & 61.84 & 0.45 & 28.89 & 81.86 & 37.07 & 62.26 \\
DeepSeek \redcross & \underline{0.12} & \underline{15.27} & 84.34 & \underline{33.24} & 27.98 & \underline{0.18} & \underline{15.80} & 83.95 & \underline{33.31} & \underline{51.68} & \underline{0.14} & \underline{15.77} & 84.09 & \underline{33.34} & \underline{40.95} \\
EuroLLM \redcross & 0.15 & 26.35 & \underline{80.43} & 35.64 & 52.25 & 0.23 & 26.26 & \underline{80.23} & 35.57 & 60.26 & 0.16 & 28.05 & \underline{77.71} & 35.31 & 62.57 \\
Gemma \graytick & \textbf{0.66} & 27.78 & \textbf{85.23} & 37.89 & 38.99 & \textbf{1.30} & \textbf{34.25} & \textbf{85.31} & \textbf{40.29} & \textbf{65.22} & \textbf{1.48} & 37.53 & \textbf{84.82} & \textbf{41.28} & 67.30 \\
LLaMA \greentick & 0.25 & 24.83 & 83.50 & 36.19 & \underline{8.85} & 1.07 & 33.08 & 82.13 & 38.76 & 63.97 & 1.43 & \textbf{39.99} & 79.26 & 40.23 & \textbf{68.91} \\
Mistral \redcross & 0.22 & 23.59 & 84.55 & 36.12 & 43.13 & 0.44 & 23.94 & 83.65 & 36.01 & 58.03 & 0.60 & 24.72 & 82.97 & 36.09 & 61.02 \\
Qwen \redcross & 0.57 & \textbf{32.25} & 84.30 & \textbf{39.04} & 58.20 & 0.76 & 33.78 & 84.11 & 39.55 & 60.40 & 1.06 & 36.60 & 84.24 & 40.64 & 64.27 \\
\bottomrule
\end{tabular}
}
\label{tab:thai}
\end{table*}

\begin{table*}[ht!]
\centering
\footnotesize 
\setlength{\tabcolsep}{2pt}
\caption{\textbf{Results on Tamil (\emph{OasisSimp-TA}) dataset}: Zero, 1- and 5-shot Results. Highest value in \textbf{bold} and lowest in \underline{underlined}. \greentick: Language included in LLM training; \graytick: Likely included (Educated Guess); \redcross: Not included.}

\resizebox{\textwidth}{!}{
\begin{tabular}{l ccccc | ccccc | ccccc}
\toprule
\multirow{3}{*}{\textbf{Model}} & \multicolumn{5}{c|}{\textbf{0 Shot}} & \multicolumn{5}{c|}{\textbf{1 Shot}} & \multicolumn{5}{c}{\textbf{5 Shot}} \\
\cmidrule(lr){2-6} \cmidrule(lr){7-11} \cmidrule(lr){12-16}
& \multicolumn{3}{c}{\textbf{SARI Comp.}} & \multirow{2}{*}{\textbf{SARI}} & \multirow{2}{*}{\textbf{F}\textsubscript{ref}} & \multicolumn{3}{c}{\textbf{SARI Comp.}} & \multirow{2}{*}{\textbf{SARI}} & \multirow{2}{*}{\textbf{F}\textsubscript{ref}} & \multicolumn{3}{c}{\textbf{SARI Comp.}} & \multirow{2}{*}{\textbf{SARI}} & \multirow{2}{*}{\textbf{F}\textsubscript{ref}} \\
\cmidrule(lr){2-4} \cmidrule(lr){7-9} \cmidrule(lr){12-14}
& \textbf{ADD} & \textbf{KEEP} & \textbf{DEL} & & & \textbf{ADD} & \textbf{KEEP} & \textbf{DEL} & & & \textbf{ADD} & \textbf{KEEP} & \textbf{DEL} & & \\
\midrule
Aya \redcross & 2.42 & 18.48 & 70.60 & 30.50 & 72.42 & 2.25 & 27.49 & 68.73 & 32.82 & 74.40 & 2.00 & 34.27 & 66.62 & 34.30 & 76.24 \\
Cmd-R \redcross & 2.78 & 26.40 & 69.28 & \textbf{32.82} & 74.14 & 2.07 & 30.78 & \underline{66.61} & 33.15 & 74.97 & 1.44 & 34.40 & 66.15 & 34.00 & 76.08 \\
DeepSeek \redcross & 0.33 & \underline{11.88} & 70.83 & \underline{27.68} & \underline{-0.67} & 0.49 & \underline{19.23} & \textbf{69.54} & \underline{29.75} & 68.27 & \underline{0.11} & \underline{8.76} & \textbf{70.56} & \underline{26.48} & \underline{3.30} \\
EuroLLM \redcross & \underline{0.19} & 24.90 & \underline{68.07} & 31.05 & 69.89 & \underline{0.32} & 26.56 & 67.05 & 31.31 & 71.21 & 0.16 & 25.05 & 67.73 & 30.98 & 65.97 \\
Gemma \graytick & \textbf{4.22} & 23.15 & \textbf{71.07} & \textbf{32.82} & \textbf{74.55} & \textbf{5.17} & \textbf{35.26} & 68.65 & \textbf{36.36} & \textbf{77.01} & \textbf{5.16} & \textbf{47.11} & \underline{65.73} & \textbf{39.34} & \textbf{79.70} \\
LLaMA \greentick & 0.75 & 19.17 & 69.88 & 29.93 & 9.71 & 1.03 & 28.81 & 66.64 & 32.16 & \underline{64.14} & 0.51 & 28.46 & 67.42 & 32.13 & 40.07 \\
Mistral \redcross & 0.57 & 16.71 & 70.29 & 29.19 & 27.62 & 1.19 & 28.34 & 67.35 & 32.29 & 72.23 & 0.89 & 32.65 & 66.40 & 33.31 & 75.04 \\
Qwen \redcross & 1.76 & \textbf{26.58} & 69.32 & 32.55 & 73.53 & 1.64 & 29.61 & 68.07 & 33.11 & 74.81 & 1.36 & 33.94 & 66.22 & 33.84 & 76.03 \\
\bottomrule
\end{tabular}
}
\label{tab:tamil}
\end{table*}

\begin{table*}[ht!]
\centering
\footnotesize
\setlength{\tabcolsep}{2pt}
\caption{\textbf{Results on Pashto (\emph{OasisSimp-PS}) dataset}: Zero, 1-, and 5-shot Results. Highest value in \textbf{bold} and lowest in \underline{underlined}.  \greentick: Language included in LLM training; \graytick: Likely included (Educated Guess); \redcross: Not included.}
\label{tab:sari_zero_pashto_triple}
\resizebox{\textwidth}{!}{
\begin{tabular}{l ccccc | ccccc | ccccc}
\toprule
\multirow{3}{*}{\textbf{Model}} & \multicolumn{5}{c|}{\textbf{0 Shot}} & \multicolumn{5}{c|}{\textbf{1 Shot}} & \multicolumn{5}{c}{\textbf{5 Shot}} \\
\cmidrule(lr){2-6} \cmidrule(lr){7-11} \cmidrule(lr){12-16}
& \multicolumn{3}{c}{\textbf{SARI Comp.}} & \multirow{2}{*}{\textbf{SARI}} & \multirow{2}{*}{\textbf{F}\textsubscript{ref}} & \multicolumn{3}{c}{\textbf{SARI Comp.}} & \multirow{2}{*}{\textbf{SARI}} & \multirow{2}{*}{\textbf{F}\textsubscript{ref}} & \multicolumn{3}{c}{\textbf{SARI Comp.}} & \multirow{2}{*}{\textbf{SARI}} & \multirow{2}{*}{\textbf{F}\textsubscript{ref}} \\
\cmidrule(lr){2-4} \cmidrule(lr){7-9} \cmidrule(lr){12-14}
& \textbf{ADD} & \textbf{KEEP} & \textbf{DEL} & & & \textbf{ADD} & \textbf{KEEP} & \textbf{DEL} & & & \textbf{ADD} & \textbf{KEEP} & \textbf{DEL} & & \\
\midrule
Aya \redcross & 0.62 & 23.98 & 67.47 & 30.69 & 49.17 & 1.08 & 45.60 & 58.62 & 35.10 & 60.83 & 1.77 & 53.81 & 47.17 & 34.25 & 68.25 \\
Cmd-R \redcross & 0.75 & 50.82 & 51.73 & 34.44 & 61.91 & 0.93 & 54.41 & 44.44 & 33.26 & 67.84 & \underline{0.70} & \textbf{56.53} & \underline{35.62} & \underline{30.95} & \textbf{70.52} \\
DeepSeek \redcross & 0.52 & 41.19 & 60.71 & 34.14 & 38.65 & 0.90 & 48.83 & 54.59 & 34.78 & 63.65 & 0.91 & 50.16 & 52.51 & 34.53 & 66.26 \\
EuroLLM \redcross & \underline{0.50} & \textbf{54.28} & \underline{44.40} & 33.06 & \textbf{67.55} & \underline{0.65} & \textbf{54.87} & \underline{43.28} & \underline{32.93} & \textbf{69.72} & 0.78 & 55.37 & 42.09 & 32.75 & 70.42 \\
Gemma \redcross & \textbf{3.84} & 25.08 & \textbf{70.78} & 33.23 & 56.95 & \textbf{4.47} & \underline{34.75} & \textbf{68.57} & 35.93 & 61.47 & \textbf{5.39} & 46.39 & \textbf{61.95} & \textbf{37.91} & 66.04 \\
LLaMA \redcross & 0.70 & \underline{18.34} & 70.28 & \underline{29.77} & \underline{-22.40} & 3.15 & 46.28 & 61.67 & \textbf{37.04} & \underline{51.15} & 1.96 & 46.53 & 58.15 & 35.55 & \underline{33.03} \\
Mistral \redcross & 0.94 & 26.36 & 68.13 & 31.81 & 47.73 & 1.42 & 41.20 & 63.04 & 35.22 & 61.31 & 1.51 & \underline{45.93} & 58.60 & 35.35 & 64.40 \\
Qwen \redcross & 2.34 & 47.48 & 58.92 & \textbf{36.25} & 58.02 & 2.81 & 49.88 & 55.34 & 36.01 & 64.76 & 2.62 & 53.79 & 48.71 & 35.04 & 65.57 \\
\bottomrule
\end{tabular}
}
\label{tab:pashto}
\end{table*}

\paragraph{Zero-Shot Performance:}
The zero-shot evaluation shows clear differences in how language and model strength affect multilingual sentence simplification. As a high-resource language, English BERTScore has less variation. This reflects the strong baseline ability of LLMs in languages that match their training data. In contrast, low- and mid-resource languages like Pashto and Thai show the largest variation in the BERTScore across models, which suggests that zero-shot is not effective for all LLMs. However, the Gemma model performs surprisingly well, achieving high BERTScore scores in Sinhala (66.77), Tamil (74.55), and Pashto (56.95). This indicates that these models transfer knowledge across languages more effectively and follow instructions better. DeepSeek and Llama's negative scores for Tamil and Pashto, respectively, highlight that some models struggle to generalize in low-resource settings.
Further analysis of SARI and its components (ADD, KEEP, DEL) gives more insight into how models behave during zero-shot simplification. The English SARI score has less variation across models, compared to other languages, meaning the models align well with the original text structure for English compared to other languages. Among all SARI components, the DEL is the highest, showing that models focus on removing unnecessary or complex information. The ADD component is the lowest, especially in non-English languages, suggesting that models find it difficult to add new or helpful information. Overall, the findings show that while Gemma handles multiple languages well and is sensitive to text structure, low-resource languages still pose challenges to all LLMs.

\paragraph{Few-shot Performance}

The few-shot evaluation (1-shot and 5-shot) shows clear and consistent improvements compared to zero-shot performance across all languages and models, especially in low-resource settings. Even one example helps the models better understand the structure and style needed for sentence simplification. Across almost all languages, 5-shot improves both SARI and \textbf{F}\textsubscript{ref} over 0-shot, especially for Gemma and Qwen. For instance, English (Gemma 39.4 to 43.6 SARI), Sinhala (Gemma 34.8 to 39.9), Thai (Gemma 37.9 to 41.3), Tamil (Gemma 32.8 to 39.3), and Pashto (Gemma 33.3 to 37.9) all show clear gains from 0 to 5 shots. This indicates that in-context examples enhance the model’s ability to balance meaning preservation with simplification. These results demonstrate that few-shot examples improve both lexical and semantic alignment in medium-resource settings. 

\paragraph{Discussion:} The SARI component analysis shows that LLMs handle content retention, deletion, and addition differently across languages. The DEL component is the most stable, with consistently high values and high scores in English (68–78) and Thai (77–85), showing that models remove unnecessary detail. Even in low-resource languages like Pashto (Gemma 70), models delete unnecessary parts for simplification. The KEEP component improves notably with few-shot learning, particularly where zero-shot transfer is weak. For example, LLaMA’s KEEP score in Thai increases from 24.83 to 39.99, demonstrating better retention and structural alignment when examples are provided. Gemma performs consistently well across languages, showing a strong ability to remove redundant material while maintaining fluency and coherence.

The ADD component remains the most variable and challenging simplification aspect, with consistently low values across all languages. English achieves modest results (5.24 - 11.91), while scores in low-resource languages remain minimal, including Sinhala (0.00–5.65), Thai (0.12–1.48), Tamil (0.11–5.16), and Pashto (0.50–5.39). This challenge is most evident in low-resource settings. For Sinhala and Pashto, except for Gemma, the ADD value is below 1 in most cases, indicating limited ability to generate new simple words. Overall, while LLMs are good at deleting unnecessary information and retaining key content, they struggle to generate new and contextually meaningful text. This limitation is especially pronounced in low-resource and morphologically complex languages for sentence simplification.

The SARI scores for English remain consistent across all models, likely because English is included in every model’s training data and benefits from extensive available resources. LLaMA shows relatively high SARI scores for Tamil and Thai, indicating that direct language inclusion during training can positively impact performance. However, it is difficult to draw firm conclusions due to the limited transparency regarding the complete set of languages and the training data size for each model. 

\section{Conclusion}
\label{sec:conclusion}
This paper introduced OasisSimp, a new multilingual dataset for sentence-level simplification, covering English, Sinhala, Thai, Tamil, and Pashto. To ensure high-quality data, all simplifications were produced by human annotators. The dataset focuses on asian languages that have been largely overlooked in existing simplification research. We evaluated the performance of eight LLMs on the OasisSimp dataset under zero-shot and few-shot learning settings. The results indicate that all models face challenges in zero-shot scenarios; however, their performance improves substantially when provided with a few examples, demonstrating the effectiveness of few-shot learning for sentence simplification. Gemma exhibited the most consistent performance across languages and conditions among the tested models.
Overall, OasisSimp represents a significant step toward broadening the scope of sentence simplification research beyond high-resource languages. This work lays the foundation for developing and evaluating more inclusive and equitable sentence simplification systems by providing high-quality, human-annotated data for less-resourced languages such as Sinhala and Pashto. We hope that OasisSimp will encourage future research on multilingual simplification and contribute to making written information more accessible to diverse global audiences.

\section{Ethical Considerations and Limitations}
\subsection{Limitations}
Language selection was purely dependent on the availability of speakers of the corresponding language. Therefore, it is not possible to carry out a systematic study with respect to language classes. The dataset size was limited, at the same time comparable to existing datasets, by the availability of annotators and/or funding. This also affected the number of simplified sentences for some languages and resulted in no human evaluation of the LLM output. There wasn't enough data to train LLMs for the sentence simplification task, and human evaluation of LLM outputs was not conducted due to resource constraints.

Due to the lack of detailed information on the languages included in training and the sizes of the data used for each LLM, it is difficult to draw concrete conclusions about language inclusion in LLMs. A more in-depth analysis of how language-family relationships and cross-linguistic transfer affect model performance would provide valuable insights, but it is beyond the scope of this work.

\subsection{Ethical Considerations}
All the complex sentences were retrieved from publicly available authentic sources (e.g., government documents, news, and Wikipedia). The simplification guidelines issued for human participants clearly stated the need to retain the original meaning of the sentence. Therefore, we do not expect any undesirable content in the simplified version of the corpus. However, the authors themselves did not individually check all the sentences for undesirable content. 
The annotators were offered co-authorship or paid at the standard rates for the country in which they reside.

\section{Acknowledgement}
This work was partially supported by the Natural Sciences and Engineering Research Council of Canada (NSERC) Discovery Grant RGPIN-2024-06887, the NSERC Discovery Launch Supplement DGECR-2024-00008, and the Digital Research Alliance of Canada (formerly Compute Canada) Grant RRG no. 5397 on "Multilingual multicultural NLP and LLMs". We also thank CoolWei AI Lab for providing GPU resources that enabled this research. This work was partially supported by the ELOQUENCE project (grant number 101070558) funded by the UKRI and the European Union. Views and opinions expressed are, however, those of the author(s) only and do not necessarily reflect those of the UKRI, European Union, or European Commission-EU. Neither the European Union nor the granting authority can be held responsible for them.

\newpage 

\section{Bibliographical References}\label{sec:reference}
\bibliographystyle{lrec2026-natbib}
\bibliography{custom}

\newpage 

\section{Appendix}

Other than the SARI and the BERTScore (\textbf{F}\textsubscript{ref}) discussed in section~\ref{sec:experimental_results}, the models are also evaluated on OasisSimp using BLEU scores \citep{papineni-etal-2002-bleu} in Table~\ref{tab:bleu_eng}, Table~\ref{tab:bleu_sin}, Table~\ref{tab:bleu_thai}, Table~\ref{tab:bleu_ta}, and Table~\ref{tab:bleu_pa}. 

Performance varies across languages: EuroLLM achieves the best results for English, Gemma performs best for Sinhala, while Cmd-R and LLaMA obtain the strongest results for Thai, depending on the shot setting. For Tamil and Pashto, models Cmd-R and Gemma achieve the best performance on Tamil, while models EuroLLM and Cmd-R perform best on Pashto, depending on the number of shots.

\begin{table}[ht!]
\centering
\caption{BLEU scores on \textbf{English (\emph{OasisSimp-EN})} dataset across different shot settings. Highest value in \textbf{bold} and lowest \underline{underlined}.}
\begin{tabular}{lccc}
\toprule
& \multicolumn{3}{c}{English}\\
\cmidrule(lr){2-4}
Model
& 0 Shot & 1 Shot & 5 Shot\\
\midrule
Aya & 18.38 & 20.59 & 21.72 
\\
Cmd-R & 21.03 & 19.58 & 20.72 
\\
DeepSeek & 17.19  & 17.27 & \underline{18.91} \\
EuroLLM & \textbf{23.07} & \textbf{23.07} & \textbf{22.79}  \\
Gemma & 15.48 & \underline{16.00} & 18.92 \\
LLaMA & 22.25 & 21.52 & 21.78 \\
Mistral & 17.40 & 19.67 & 20.63 \\
Qwen & \underline{11.55} & 17.21 & 19.09\\
\bottomrule
\end{tabular}

\label{tab:bleu_eng}
\end{table}

\begin{table}[ht!]
\centering
\caption{BLEU scores on\textbf{ Sinhala(\emph{OasisSimp-SI})} dataset across different shot settings. Highest value in \textbf{bold} and lowest \underline{underlined}.}
\begin{tabular}{lccc}
\toprule
& \multicolumn{3}{c}{Sinhala}\\
\cmidrule(lr){2-4}
Model
& 0 Shot & 1 Shot & 5 Shot\\
\midrule
Aya & 7.25 & 9.03 & 10.80 
\\
Cmd-R  & 11.86 & 11.90 & 11.63 
\\
DeepSeek& 8.04 & 7.12 & 3.08 \\
EuroLLM & \underline{2.24} & \underline{2.19} & \underline{0.00} \\
Gemma & \textbf{14.38} & \textbf{22.36} & \textbf{31.82} \\
LLaMA & 6.45 & 6.77 & 7.34\\
Mistral & 2.29 & 4.55 & 7.72  \\
Qwen  & 12.28 & 12.64 & 13.02 \\
\bottomrule
\end{tabular}

\label{tab:bleu_sin}
\end{table}

\begin{table}[ht!]
\centering
\caption{BLEU scores on \textbf{Thai(\emph{OasisSimp-TH})} dataset across different shot settings. Highest value in \textbf{bold} and lowest \underline{underlined}.}
\begin{tabular}{lccc}
\toprule
& \multicolumn{3}{c}{Thai}\\
\cmidrule(lr){2-4}
Model
& 0 Shot & 1 Shot & 5 Shot\\
\midrule
Aya &7.70 & 9.65 & 11.58 
\\
Cmd-R &  \textbf{13.08} & 12.42 & 13.84
\\
DeepSeek & 2.24 & 5.99 & \underline{5.71} \\
EuroLLM & 11.76 & 13.94 & 16.73 \\
Gemma & 3.67 & 11.20 & 14.43 \\
LLaMA & \underline{2.16} & \textbf{14.71} & \textbf{20.36} \\
Mistral & 4.54 & 8.52 & 9.79 \\
Qwen & 5.35 & \underline{5.55} & 9.16\\
\bottomrule
\end{tabular}

\label{tab:bleu_thai}
\end{table}

\begin{table}[ht!]
\centering
\caption{BLEU scores on \textbf{Tamil(\emph{OasisSimp-TA})} dataset across different shot settings. Highest value in \textbf{bold} and lowest \underline{underlined}. }
\begin{tabular}{lccc}
\toprule
& \multicolumn{3}{c}{Tamil}\\
\cmidrule(lr){2-4}
Model
& 0 Shot & 1 Shot & 5 Shot\\
\midrule
Aya & 11.26 & 20.69 & 29.97 
\\
Cmd-R & \textbf{19.36} & \textbf{26.17} & 30.58
\\
DeepSeek & \underline{3.67} & \underline{10.11} & \underline{1.48} \\
EuroLLM & 17.04 & 18.98 & 15.60 \\
Gemma & 9.62 & 21.36 & \textbf{36.71}
\\
LLaMA & 7.94 & 23.17 & 22.21 
\\
Mistral & 7.38 & 22.21 & 27.66
\\
Qwen &  18.77 & 23.30 & 29.67 
\\
\bottomrule
\end{tabular}

\label{tab:bleu_ta}
\end{table}

\begin{table}[ht!]
\centering
\caption{BLEU scores on \textbf{Pashto(\emph{OasisSimp-PS})} dataset across different shot settings. Highest value in \textbf{bold} and lowest \underline{underlined}. }
\begin{tabular}{lccc}
\toprule
& \multicolumn{3}{c}{Pashto} \\
\cmidrule(lr){2-4}
Model
& 0 Shot & 1 Shot & 5 Shot\\
\midrule
Aya & 7.42 & 22.04 & 32.73
\\
Cmd-R & 28.39 & 33.99 & \textbf{38.49}
\\
DeepSeek & 18.37 & 25.15 & 26.92\\
EuroLLM & \textbf{33.74} & \textbf{34.65} & 35.54\\
Gemma & 7.22 & \underline{12.84} & 22.65
\\
LLaMA & \underline{5.28} & 22.09 & 23.30
\\
Mistral & 7.32 & 16.11 & \underline{20.95}
\\
Qwen & 21.03 & 27.00 & 30.01
\\
\bottomrule
\end{tabular}

\label{tab:bleu_pa}
\end{table}

\end{document}